\documentclass[letterpaper, 10 pt, conference]{ieeeconf}
\pdfoutput=1

\IEEEoverridecommandlockouts
\usepackage{times}
\usepackage{multicol}
\usepackage{algorithm}
\usepackage{algpseudocode}
\usepackage{amsmath}
\usepackage{amssymb}
\usepackage{booktabs}
\usepackage{graphicx}
\usepackage{multirow}
\usepackage{xcolor}
\usepackage[most]{tcolorbox}
\let\labelindent\relax
\usepackage{enumitem}
\usepackage{flushend}
\usepackage[
    bookmarks=true,
    colorlinks=true,
    urlcolor=blue,
    linkcolor=blue,
    citecolor=blue  
]{hyperref}

\newtcolorbox{PromptBox}[1]{
    colback=gray!5, 
    colframe=gray!50, 
    title=#1,
    fonttitle=\small\bfseries,
    fontupper=\small,
    arc=2pt,
    boxrule=0.5pt,
    left=5pt,
    right=5pt,
    top=5pt,
    bottom=5pt
}

\pdfoutput=1
\title{\LARGE \bf
Robot Planning and Situation Handling \\with Active Perception
}

\author{Austine Oloo$^1$, Zainab Altaweel$^1$, Yohei Hayamizu$^1$, Peiqi Liu$^2$, Yan Ding$^1$, Saeid Amiri$^1$, Hao Yang$^3$\\
Andy Kaminski$^3$, Chad Esselink$^3$, Chris Paxton$^4$, Xiaohan Zhang$^1$, Shiqi Zhang$^1$
\thanks{$^1$SUNY Binghamton, $^2$CMU, $^3$Ford Research, $^4$Agility Robotics}
}

\begin{document}

\maketitle
\thispagestyle{empty}
\pagestyle{empty}

\begin{abstract}
Current robots are capable of computing plans to accomplish complex tasks. 
However, real-world environments are inherently open and dynamic, and unforeseen situations frequently arise during plan execution, such as jamming doors and fallen objects on the floor. 
These situations may result from the robot's own action failures or from external disturbances, such as human activities. 
Detecting and handling such execution-time situations remains a significant challenge, limiting those robots' ability to achieve long-term autonomy. 
In this paper, we develop a planning and situation-handling framework, called VAP-TAMP, that enables robots to actively perceive and address unforeseen situations during plan execution. 
VAP-TAMP leverages action knowledge to strategically prompt vision-language models for active view selection and situation assessment, while constructing and reasoning over scene graphs for integrated task and motion planning. 
We evaluated VAP-TAMP using service tasks in simulation and on a mobile manipulation platform. 
Project page: \href{https://vap-tamp.github.io/vap-tamp/}{vap-tamp.github.io/vap-tamp}
\end{abstract}

\IEEEpeerreviewmaketitle

\section{Introduction}
\label{sec:introduction}
Robots need planning capabilities to sequence actions to achieve complex goals. 
Task planning is on high-level symbolic actions; and motion planning methods generate continuous trajectories for realizing those discrete actions in geometric spaces. 
Integrated task and motion planning~(TAMP) algorithms compute task-motion plans that are in the form of multiple motion trajectories for achieving complex goals~\cite{garrett2021integrated,zhao2024survey}. 
While existing TAMP methods well support plan generation, unforeseen situations frequently get the robots stuck during plan execution. 

Prior work has explored several directions to address unforeseen situations in open-world planning. 
Example open-world planning ideas include novel knowledge representations for explaining and addressing task failures~\cite{hanheide2017robot}, leveraging large language models (LLMs) to dynamically augment action knowledge for situation handling~\cite{ding2023integrating}, and triggering human intervention when the robot is unsure about solutions~\cite{ren2023robots}. 
These methods implicitly assumed that situations are \emph{fully observable}, and hence directly focused on addressing those situations. 
In the real world, situations are only \emph{partially observable}, and robots will have to rely on its on-board, unreliable sensors to perceive the situations before tackling them. 

\begin{figure}[t]
    \centering
    \vspace{.5em}
    \includegraphics[width=0.9\columnwidth]{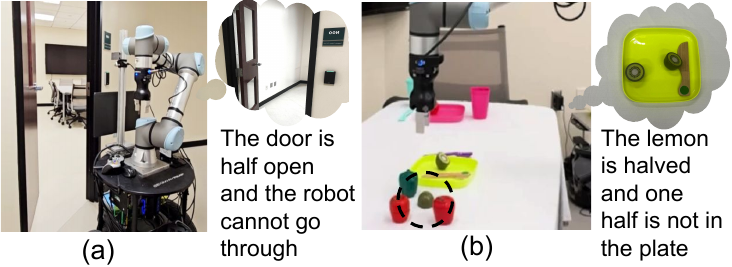}
    \vspace{-.6em}
    \caption{Two examples of unforeseen situations during action execution. 
    On the left, the robot attempted to navigate through a doorway to reach the room. It was expected that the door would be fully open and passable (envisioned in the ``cloud''), while the door was only half-open. 
    On the right shows the robot cutting a lemon. It was expected that both lemon halves would remain in the plate, while one half had fallen outside.}
    \label{fig:Tasks}
\end{figure}

In this paper, we present \textbf{VAP-TAMP}, a novel TAMP framework that performs situation handling through \underline{V}LM-based \underline{A}ctive \underline{P}erception. 
VAP-TAMP dynamically maintains a scene graph as the symbolic state representation and leverages vision-language models (VLMs)~\cite{team2023gemini} for plan monitoring and active perception. 
The novelty of VAP-TAMP lies in the interaction between VLMs and action knowledge. 
VAP-TAMP actively prompts VLMs using the predicates of the current action's preconditions and effects in action knowledge, while leveraging the VLM outputs to update the action knowledge for situation handling. 
At the motion level, VAP-TAMP prompts VLMs for active viewpoint selection towards maximizing information gain. 
At the task level, VLMs are used for visually evaluating whether preconditions and effects are satisfied before and after each action for plan monitoring. 
Action knowledge is encoded in PDDL~\cite{haslum2019introduction} in this work. 
This integration of VLMs and PDDL-based action knowledge enables the robots to actively detect and address execution-time situations, enabling robust plan generation and execution in open worlds. 

VAP-TAMP has been extensively evaluated in simulation and real-world settings, where the robot is challenged by novel situations, as illustrated in Fig~\ref{fig:Tasks}. 
In our real-world experiments with 100 service task trials on a mobile manipulator, VAP-TAMP achieved 88\% task success rate, whereas baselines OK-Robot~\cite{Liu_2024} and COWP~\cite{ding2023integrating} achieved 76\% and 69\% respectively. 

\section{Related Work}
\label{sec:related_work}

\noindent
\textbf{Task and Motion Planning (TAMP).}
Early work on belief-space TAMP addressed partial observability but requires known uncertainty models~\cite{kaelbling2013integrated}.
Subsequent approaches like PDDLStream and FFRob achieve impressive performance but assume fully observable, static environments with perfect state estimation~\cite{garrett2020pddlstream, garrett2018ffrob}.
Recent work addresses visual grounding in TAMP. 
One example is LLM-GROP that grounds predicates in visual observations using language models~\cite{zhang2025llm}. 
For goal specification, L3M+P extends this with external knowledge graphs for lifelong planning~\cite{agarwal2025l3m+}.
Recent work such as TAMPURA~\cite{curtis2024partially} extends TAMP to reason under uncertainty and risk.
Another line of work guides TAMP using machine learning and reinforcement learning to improve planning efficiency in complex manipulation tasks~\cite{chitnis2016guided} .
Recent articles reviewed TAMP literature~\cite{garrett2021integrated,zhao2024survey}. 
Our work assumes partial observability of the world states (and situations), and leverages VLMs for active perception to detect and resolve situations at plan execution time.

\vspace{.5em}
\noindent 
\textbf{Vision-Language Models for Robot Planning.}
VLMs have transformed robotic perception through open-vocabulary scene understanding~\cite{radford2021learning, Liu_2024}. SayCan grounds language instructions via learned affordances~\cite{ahn2022can}, Code as Policies generates executable robot code~\cite{liang2022code}, and PaLM-E integrates visual tokens for embodied reasoning~\cite{driess2023palm}. VoxPoser synthesizes 3D value maps from VLM outputs~\cite{huang2023voxposer}.

However, these systems treat VLM outputs as infallible oracles which are queried once, trusted implicitly, never verified. VLMs exhibit systematic failures under occlusion and suboptimal viewpoints~\cite{rahmanzadehgervi2024vision}, yet provide no calibrated uncertainty signal~\cite{lin2023generating}. 
By comparison, VAP-TAMP (ours) transforms VLM uncertainty into an actionable signal that triggers deliberate information gathering.

\vspace{.5em}
\noindent 
\textbf{Active Perception in Robotics.}
There is a long history of research on active perception in robotics~\cite{bajcsy1988active, bajcsy2018revisiting}. 
For example, next-best-view (NBV) algorithms maximize geometric coverage~\cite{kriegel2011surface, isler2016information}. 
NBV methods optimize over geometric primitives and cannot determine which viewpoints would resolve VLM uncertainty about semantic predicates such as containment or spatial relationships. 
Interactive perception methods have been developed to physically interact with objects to reveal occluded surfaces~\cite{bohg2017interactive,zhang2021planning}. 
Different from those methods, VAP-TAMP prompts VLMs with its current state and action information to actively select viewpoints and verify predicate values towards successful plan execution. 

\vspace{.5em}
\noindent 
\textbf{Scene Graphs for Robot Reasoning.}
Scene graphs provide structured representations for robot reasoning~\cite{rosinol2021kimera, hughes2022hydra, amiri2022reasoning}.
ConceptGraphs grounds open-vocabulary concepts using VLM features~\cite{gu2024conceptgraphs}, and OK-Robot combines such representations with learned manipulation primitives~\cite{Liu_2024}. 
VAP-TAMP maintains a dynamic scene graph, where the process is driven by its percepts, actions and states. 
In this work, scene graphs, as a representation of world state, are integrated into TAMP, VLM prompting and situation handling. 

\vspace{.5em}
\noindent 
\textbf{Failure Detection and Recovery.}
Robust execution requires detecting failures and generating recovery strategies~\cite{ingrand2017deliberation}. Classical approaches monitor preconditions and trigger replanning but assume accurate state estimation~\cite{brenner2006continual}. SuccessVQA verifies action outcomes but without quantifying confidence~\cite{du2023vision}. Inner Monologue incorporates multi-modal feedback but cannot actively resolve ambiguity~\cite{huang2022inner}. SayPlan uses 3D scene graphs for grounded replanning but inherits static representation limitations~\cite{rana2023sayplan}. 
By comparison, VAP-TAMP (ours) diagnoses whether discrepancies stem from perception error (correctable through active perception), world state change (requiring plan adaptation), or execution failure (requiring recovery).

\section{Problem Formulation}
\label{sec:problem}
We consider a mobile manipulator tasked with achieving goals specified in natural language in open world indoor environments. The system receives the following inputs. 
\begin{itemize}
    \item RGB-D observations $o_t \in \mathcal{I}$ from onboard cameras at each timestep $t$.
    \item Robot pose $\xi_t \in SE(3)$ from localization.
    \item Natural language goal $g_{\text{NL}}$ (e.g., ``store the firewood in the basket'').
    \item Planning domain $\mathcal{D} = \langle \mathcal{T}, \mathcal{P}, \mathcal{A} \rangle$ specifying object types $\mathcal{T}$, predicates $\mathcal{P}$, and actions $\mathcal{A}$ with preconditions and effects defined over $\mathcal{P}$. For example, $\mathcal{P}$ includes predicate symbols such as \texttt{on}, \texttt{holding}, and \texttt{hand\_empty}; there are preconditions $\{\texttt{on(obj, surface)}, \texttt{hand\_empty}\}$ and effects $\{\texttt{holding(obj)}, \neg\texttt{on(obj, surface)}\}$ for action \texttt{pick(obj, surface)}.
\end{itemize}

We assume access to a library of parameterized motion primitives (such as navigate, pick, place, and push) with associated motion planners. The system determines which primitives to invoke and with what parameters.
We further assume the robot can localize within a known map and that objects of interest are detectable, though their symbolic properties must be inferred from visual observations.
The objective is to generate and execute a sequence of actions that achieves the goal, maximizing task success rate:
\begin{equation}
    \max \left( \mathbb{P}\left( s_T \models g \right) \right)
\end{equation}
where $\mathbb{P}(\cdot)$ denotes probability, $s_T$ is the final symbolic state, and $g$ is the goal condition derived from $g_{\text{NL}}$.

Open world settings are fundamentally characterized by environmental dynamism, e.g., 
objects are displaced by humans, access paths become blocked, and task-relevant conditions shift. 
This dynamism manifests in three interrelated challenges. 
\textbf{First}, perception is unreliable. 
Predicate evaluation from visual observations may produce errors due to occlusion, clutter, or suboptimal viewing angles, yet provides no confidence signal to indicate when estimates are wrong. 
\textbf{Second}, the environment changes between planning and execution, causing plans computed from initial observations to become invalid. 
\textbf{Third}, actuation is imperfect, e.g., manipulation actions may fail to achieve their intended effects, with grasps slipping or placements missing their targets. 
When classical systems encounter these failures, they either halt or require human intervention.

We aim to develop robot systems that autonomously evaluate perceptual uncertainty, actively gather disambiguating observations when the current view is insufficient, and recover from execution failures through replanning.

\section{Methodology}
\label{sec:methodology}
VAP-TAMP is a TAMP framework that performs situation handling through VLM-based active perception. 
The system maintains a scene graph as symbolic state representation, leverages action knowledge to prompt VLMs for situation assessment, and actively selects new viewpoints when the current view is insufficient for reliable evaluation.

\begin{figure*}[t]
    \centering
    \hspace{5em}   \includegraphics[width=0.95\textwidth]{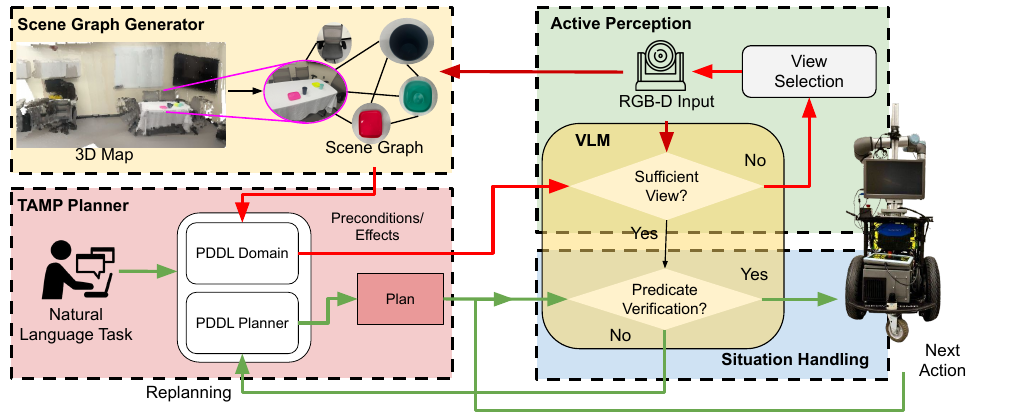}
    \vspace{-.8em}
    \caption{\textbf{VAP-TAMP System Overview.} Given RGB-D observations and a goal description in natural language, 
    VAP-TAMP builds a 3D point cloud (map), extracts an instance memory and scene graph (Section~\ref{sec:scene_graph}), and translates to PDDL for planning. 
    During execution, preconditions and effects are verified before and after each action (Section~\ref{sec:situation_handling}), with uncertainty evaluation triggering active perception when needed (Section~\ref{sec:active_perception}). 
    Verification failures at either stage constitute situations that prompt state correction and replanning.
    }
    \label{fig:overview}
\end{figure*}

Fig.~\ref{fig:overview} illustrates the architecture.
Given a natural language goal $g_{\text{NL}}$, the system parses it into a symbolic goal specification $g$, generates an action sequence $\rho$ using a PDDL planner, and executes each action with active perception for precondition and effect verification. Three modules support robust TAMP execution: 1) \textit{scene graph generator} constructs and maintains the symbolic state representation required for planning, 2) \textit{active perception module} grounds predicates using VLMs to verify action preconditions and effects, and 3) \textit{situation handling} detects execution failures and triggers replanning when the world state diverges from expectations.

Algorithm~\ref{alg:vap_tamp} presents the execution procedure. The framework operates over the planning domain $\mathcal{D}$ defined in Section~\ref{sec:problem}. Each action $a \in \mathcal{A}$ has preconditions $\text{Pre}(a)$ that must hold before execution and effects $\text{Eff}^+(a)$, $\text{Eff}^-(a)$ specifying predicates added or removed after execution.

The system first constructs a scene graph $\mathcal{G} = (\mathcal{O}, \mathcal{R})$ from RGB-D observations, where $\mathcal{O}$ is the set of detected objects and $\mathcal{R}$ is the set of grounded predicates encoding spatial relationships. A VLM parses the natural language goal $g_{\text{NL}}$ into a PDDL goal specification $g$ using the scene graph and domain knowledge. The PDDL planner then generates an action sequence $\rho = [a_0, \ldots, a_{n-1}]$ that transforms the initial state into one satisfying $g$.

For each action $a$ in the plan, execution proceeds through a verify-execute-verify cycle. The robot navigates to the task-relevant area, captures an observation, and verifies that all preconditions $\text{Pre}(a)$ hold using VLM-based predicate verification. If verification succeeds, the robot executes the action, updates the scene graph with expected effects, and verifies that intended effects were achieved. If any verification fails, the scene graph is corrected based on observation and the PDDL planner generates a recovery plan from the updated state. This tight coupling between symbolic planning and perceptual verification ensures that TAMP execution remains grounded in physical reality.

\begin{algorithm}[t]
\caption{VAP-TAMP: Closed-Loop TAMP Execution}
\label{alg:vap_tamp}
\begin{algorithmic}[1]
\footnotesize
\Require Goal $g_{\text{NL}}$, domain $\mathcal{D}$, VLM $V$
\Ensure Task success or failure
\State $\mathcal{G} \leftarrow$ BuildSceneGraph() \Comment{Sec.~\ref{sec:scene_graph}}
\State $g \leftarrow$ ParseGoal($g_{\text{NL}}$, $\mathcal{G}$, $\mathcal{D}$) \Comment{VLM-based}
\State $\rho \leftarrow$ Plan($\mathcal{G}$, $g$, $\mathcal{D}$) \Comment{PDDL planner}
\While{$\rho \neq \emptyset$}
    \State $a \leftarrow \rho$.pop(); Navigate($a$, $\mathcal{G}$); $o \leftarrow$ Observe()
    \For{$p \in \text{Pre}(a)$} \Comment{Precondition verification, Sec.~\ref{sec:situation_handling}}
        \State $v_p \leftarrow$ \Call{VerifyPredicate}{$o$, $p$, $V$, $\mathcal{G}$} \Comment{Sec.~\ref{sec:active_perception}}
        \If{$v_p = \text{False}$}
            \State $\mathcal{G} \leftarrow$ UpdateGraph($p$, $v_p$, $\mathcal{G}$)
            \State $\rho \leftarrow$ Plan($\mathcal{G}$, $g$, $\mathcal{D}$); \textbf{continue}
        \EndIf
    \EndFor
    \State Execute($a$); $\mathcal{G} \leftarrow$ ApplyEffects($a$, $\mathcal{G}$)
    \State $o \leftarrow$ Observe()
    \For{$p \in \text{Eff}^+(a)$} \Comment{Effect verification, Sec.~\ref{sec:situation_handling}}
        \State $v_p \leftarrow$ \Call{VerifyPredicate}{$o$, $p$, $V$, $\mathcal{G}$}
        \If{$v_p = \text{False}$}
            \State $\mathcal{G} \leftarrow$ UpdateGraph($p$, $v_p$, $\mathcal{G}$)
            \State $\rho \leftarrow$ Plan($\mathcal{G}$, $g$, $\mathcal{D}$); \textbf{break}
        \EndIf
    \EndFor
\EndWhile
\State \Return $\mathcal{G} \models g$
\end{algorithmic}
\end{algorithm}

\subsection{Scene Graph Generator}
\label{sec:scene_graph}

The scene graph serves as the symbolic state representation for PDDL planning, bridging perceptual observations and the discrete state space over which the planner searches. 

\subsubsection{Construction}

The system builds an initial scene graph through RGB-D exploration. RGB-D observations collected during exploration are aggregated into a point cloud map $\mathcal{M}$, and instance segmentation partitions $\mathcal{M}$ into object instances forming $\mathcal{O}$. Each object $obj_i \in \mathcal{O}$ is stored with its identifier, 3D centroid $c_i \in \mathbb{R}^3$, bounding box $b_i \in \mathbb{R}^6$, and CLIP embedding $e_i$ for open-vocabulary retrieval.

From the object set, the relationship set $\mathcal{R}$ is constructed through geometric reasoning. For each object pair $(obj_i, obj_j)$, spatial predicates are evaluated:
\begin{itemize}
    \item $\texttt{on}(obj_i, obj_j)$ holds if $obj_i$'s centroid lies above $obj_j$'s support surface within a threshold.
    \item $\texttt{inside}(obj_i, obj_j)$ holds if $obj_i$'s bounding box is contained within $obj_j$'s volume.
    \item $\texttt{near}(obj_i, obj_j)$ holds if the distance between centroids is below a threshold.
\end{itemize}

\begin{algorithm}[t]
\caption{VerifyPredicate: Active Perception for Predicate Verification}
\label{alg:active_perception}
\begin{algorithmic}[1]
\small
\Require Observation $o$, predicate $p$, VLM $V$, graph $\mathcal{G}$, budget $K$
\Ensure Predicate value $v_p \in \{\text{True}, \text{False}\}$
\For{$k = 0$ \textbf{to} $K$}
    \State $\mathcal{Q}_p \leftarrow$ GenerateParaphrases($p$) \Comment{$N$ variants}
    \State $\mathbf{y} \leftarrow [V(o, q) \text{ for } q \in \mathcal{Q}_p]$ \Comment{Query VLM}
    \State $v_p \leftarrow \mathbf{1}[\text{mean}(\mathbf{y}) > 0.5]$ \Comment{Majority vote}
    \If{responses consistent} \Comment{High agreement}
        \State $s \leftarrow$ ConfirmSufficiency($o$, $p$, $V$)
        \If{$s = \text{True}$} \Return $v_p$ \EndIf
    \EndIf
    \State $d \leftarrow$ SuggestViewpoint($o$, $p$, $V$) \Comment{VLM-guided}
    \State Navigate($d$); $o \leftarrow$ Observe()
    \State $\mathcal{G} \leftarrow$ UpdateFromObservation($o$, $\mathcal{G}$)
\EndFor
\State \Return $v_p$ \Comment{Budget exhausted}
\end{algorithmic}
\end{algorithm}

\subsubsection{Maintenance}

The scene graph must remain synchronized with physical reality as the robot manipulates the environment. Updates occur through two mechanisms:

\textit{Action-driven updates} apply immediately after action execution. Given action $a$ with effects $\text{Eff}^+(a)$ and $\text{Eff}^-(a)$, the scene graph is updated by removing predicates in $\text{Eff}^-(a)$ and adding predicates in $\text{Eff}^+(a)$. For example, executing \texttt{pick(cup, table)} removes $\texttt{on(cup, table)}$ and adds $\texttt{holding(robot, cup)}$.

\textit{Observation-driven updates} occur when predicate verification detects a discrepancy between the scene graph and physical reality. If the verified value differs from the recorded value, the scene graph is corrected accordingly. This correction triggers replanning from the updated state, ensuring the PDDL planner always operates on an accurate world model.

\subsection{Active Perception}
\label{sec:active_perception}

The robot performs active perception to support situation handling in plan execution. In particular, we leverage VLMs and domain knowledge (action preconditions and effects) for predicate verification and viewpoint selection.

The active perception module takes an observation $o$, a grounded predicate $p$ to verify (e.g., $p = \texttt{on(cup, table)}$), and the current scene graph $\mathcal{G}$. It returns a truth value $v_p \in \{\text{True}, \text{False}\}$ and a potentially updated scene graph. Algorithm~\ref{alg:active_perception} presents the procedure.

\subsubsection{Predicate Verification}

Given predicate $p$ and observation $o$, the system queries the VLM to determine whether $p$ holds. To improve reliability, we generate a set of $N$ semantically equivalent natural language questions from the predicate. For predicate $p = \texttt{on(cup, table)}$, variants include ``Is the cup on the table?'', ``Is the cup resting on the table surface?'', and ``Is the cup positioned on top of the table?''

For each query $q_i$, the VLM returns a binary response $y_i \in \{0, 1\}$ indicating whether the predicate holds. From the $N$ responses, we compute the final predicate value by majority vote:
\begin{equation}
    v_p = \mathbf{1}\left[\frac{1}{N}\sum_{i=1}^{N} y_i > 0.5\right]
\end{equation}
where $\mathbf{1}[\cdot]$ is the indicator function returning True if the condition holds.

When responses are consistent, we query the VLM to confirm whether the current view provides sufficient information. If the VLM confirms sufficiency, the predicate value $v_p$ is accepted. Otherwise, viewpoint selection is triggered.

\begin{figure}[t]
    \centering
    \vspace{-.1em}
    \includegraphics[width=\columnwidth]{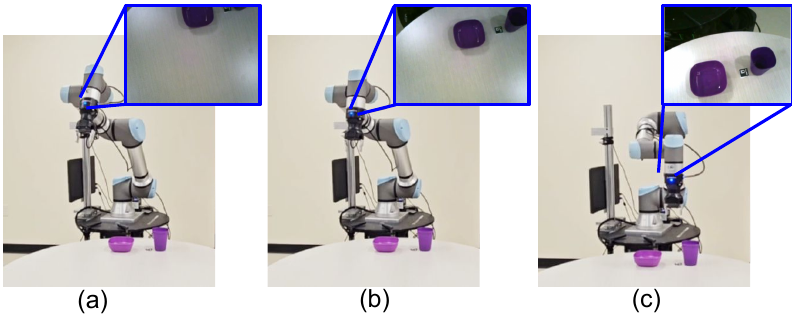}
    \vspace{-2em}
    \caption{Active perception resolving visual ambiguity during predicate verification for a \texttt{grasp} action. In (a), the robot captures an initial observation and queries the VLM with $N$ semantically equivalent paraphrases of the target predicate; inconsistent responses signal 
    perceptual uncertainty. In (b), the VLM confirms the view is insufficient and suggests a better viewing direction, prompting the robot to actively perceive the scene. In (c), the inset shows the improved close-up view from which the VLM now returns consistent responses across all paraphrases, enabling confident predicate verification, scene graph update, and plan continuation.}
    \label{fig:active_perception}
\end{figure}

\subsubsection{Viewpoint Selection}

When predicate verification cannot be completed confidently from the current viewpoint, the system queries the VLM to suggest another viewing position, as illustrated in Fig.~\ref{fig:active_perception}. 

The VLM analyzes the current observation and target predicate, returning a direction $d \in \{\text{left}, \text{right}, \text{closer}, \text{above}, \ldots\}$ that would provide potentially clearer visual evidence. The robot navigates in the suggested direction and captures new observations. 
The verification process repeats with the new observation until the VLM confirms sufficient information or a viewpoint budget $K$ is exhausted.

\begin{figure}[t]
    \centering
    \includegraphics[width=\columnwidth]{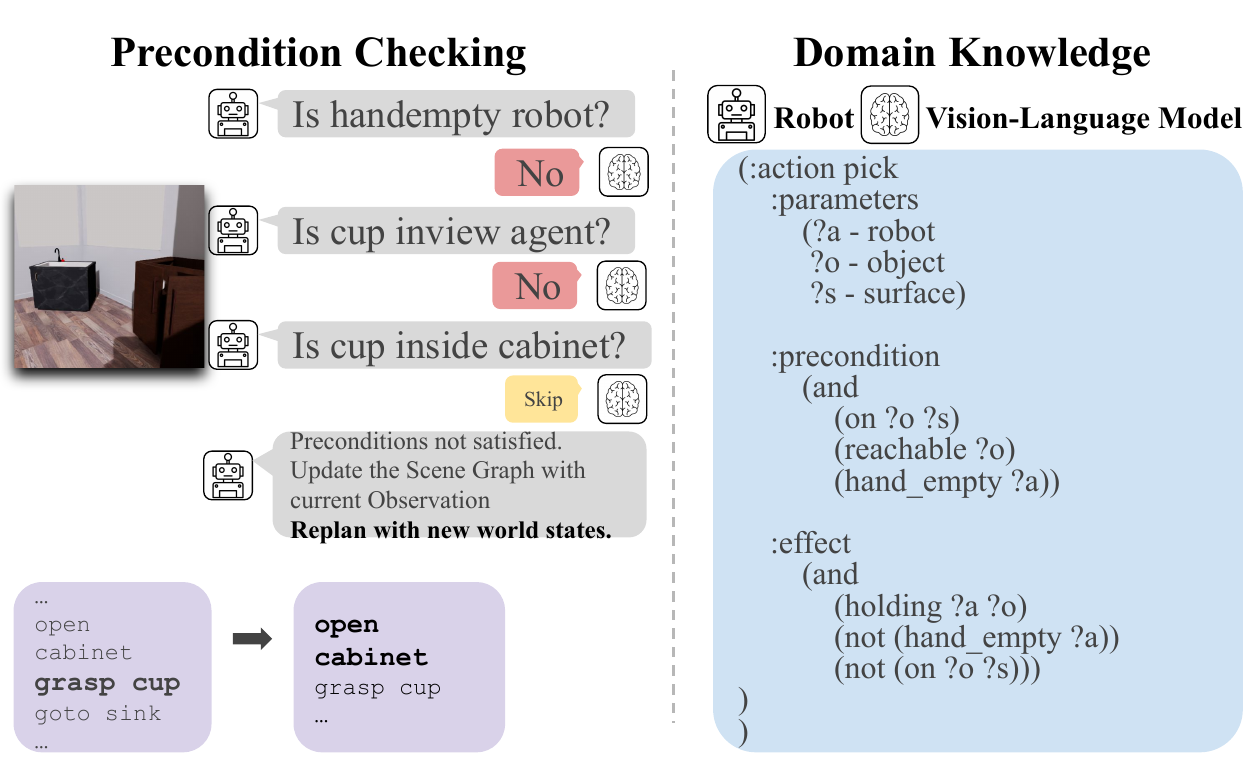}
    \vspace{-1.7em}
    \caption{VAP-TAMP integrates robot perception with domain knowledge by formulating the verifications of action preconditions and effects as visual question answering (VQA) queries. 
    The left shows an example of precondition verification, where VAP-TAMP verifies the action preconditions one at a time. 
    The precondition predicates are dynamically extracted from the PDDL-based description of the current action. 
    }
    \label{fig:situation_handling}
\end{figure}

\subsection{Situation Handling}
\label{sec:situation_handling}

The situation handler detects execution failures and triggers PDDL replanning, enabling recovery from unexpected world states as shown in Fig.~\ref{fig:situation_handling}. We define a \textit{situation} as an unexpected world state that prevents task completion using a plan that would normally succeed. Situations arise from environmental changes (objects displaced by humans), perceptual errors (incorrect initial state), and execution failures (failed grasps).

\subsubsection{Precondition Verification}

Before executing action $a$, the situation handler verifies that all preconditions $\text{Pre}(a)$ hold. For action $a = \texttt{pick(cup, table)}$, preconditions include $\texttt{on(cup, table)}$, $\texttt{reachable(cup)}$, and $\texttt{hand\_empty}$. Each predicate $p \in \text{Pre}(a)$ is verified using the active perception module.

When a precondition evaluates to False, the system detects a situation. The scene graph is corrected accordingly, and the PDDL planner generates a new plan from the corrected state.

\subsubsection{Effect Verification}

After executing action $a$, the situation handler verifies that intended effects $\text{Eff}^+(a)$ were achieved. For $a = \texttt{pick(cup, table)}$, the expected effect is $\texttt{holding(robot, cup)}$ becoming true. Each effect predicate is verified using the active perception module.

When an expected effect evaluates to False, an execution failure is detected. The scene graph is corrected based on observation rather than assumed effects, and PDDL replanning proceeds. This closed-loop verification ensures the symbolic state remains grounded in physical reality despite imperfect actuation, enabling robust TAMP execution in open-world environments.

\begin{table}[t]
    \centering
    \footnotesize
    \caption{Actions, preconditions, and situations (unexpected outcomes) that VAP-TAMP must detect and recover from.}
    \vspace{-0.3em}
    \begin{tabular}{p{.7cm}p{2.1cm}p{4.1cm}}
        \toprule
        \textbf{Action} & \textbf{Preconditions} & \textbf{Situations} \\
        \midrule
        find & Object and agent in same room & Object not inview after navigation; no free space near object; held object drops during navigation \\
        \midrule
        grasp & Object inview; hand empty & Grasp fails, object unchanged; grasp fails, object drops nearby \\
        \midrule
        placein & Object inhand; receptacle inview; receptacle open & Place fails, object remains inhand; place fails, object drops nearby \\
        \midrule
        placeon & Object inhand; receptacle inview & Place fails, object remains inhand; place fails, object drops nearby \\
        \midrule
        open & Object inview & Open fails, object remains closed \\
        \midrule
        close & Object inview & Close fails, object remains open \\
        \midrule
        turnon & Object inview & Turn on fails, object remains off \\
        \midrule
        cut & Object inview; knife inhand & Object not cut, knife inhand; object not cut, knife drops nearby \\
        \bottomrule
    \end{tabular}
    \vspace{-1.5em}
    \label{tab:situations}
\end{table}

\section{Experiments}
\label{sec:experiments}

We evaluate VAP-TAMP in both real-world and simulation settings. Our experiments test two hypotheses: (H1) VAP-TAMP achieves higher task success rate than competitive baselines across diverse mobile manipulation tasks; (H2) VLM-guided active perception for VAP-TAMP resolves visual ambiguity more efficiently than greedy perception, enabling reliable predicate verification with fewer viewpoints.

\begin{figure}[t]
    \centering
    \vspace{.5em}
    \includegraphics[width=\columnwidth]{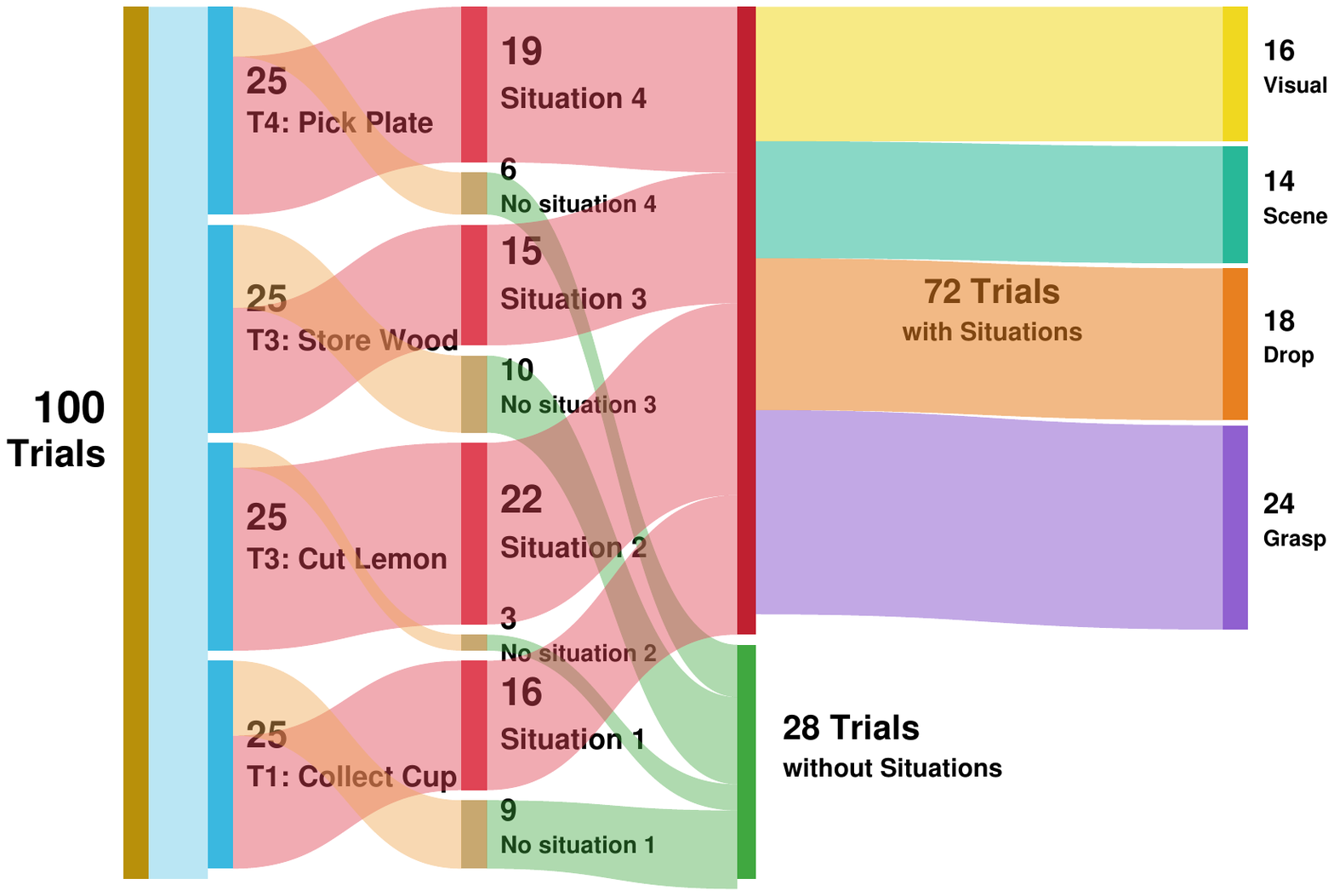}
    \vspace{-2.7em}
    \caption{Distribution of situations across tasks. Flows connect tasks (left) to situation (right), with occurrence count.}
    \vspace{-1em}
    \label{fig:situation_distribution}
\end{figure}

\subsection{Real-World Experiments}
\label{sec:real_world}

\subsubsection{Setup}

We use a mobile manipulator comprising a Segway RMP base, UR5e arm, Robotiq 2F-85 gripper, and wrist-mounted Intel RealSense D435i RGB-D camera. VLM queries use Gemini Vision~\cite{team2023gemini}. Motion primitives (navigate, pick, place, push) are implemented using MoveIt~\cite{coleman2014reducing} for arm planning.

We evaluate on four household tasks in three rooms with 25 trials each (100 total). \textbf{T1: Collect Cup} requires picking up a cup from one room and taking it to another. \textbf{T2: Cut Lemon} requires locating a lemon, finding a knife, and cutting the lemon. \textbf{T3: Store Firewood} requires collecting wooden pieces and placing them into a storage room. \textbf{T4: Pick Plate} requires retrieving a plate from a cluttered table and placing it on a shelf.

We compare against three baselines. \textbf{OK-Robot}~\cite{Liu_2024} is a state-of-the-art open-vocabulary mobile manipulation system that combines CLIP-based object retrieval with VLM-based situation handling, but without active perception. \textbf{COWP}~\cite{ding2023integrating} uses LLM commonsense to dynamically augment action knowledge for situation handling but supports task planning only (no motion planning). \textbf{Closed World} is a classical TAMP baseline with no situation handling that assumes perfect state estimation.

Table~\ref{tab:situations} shows the situations that can occur during task execution. Each action has preconditions that must hold before execution and effects that should hold after. Situations arise when preconditions are violated unexpectedly or when effects fail to materialize. These situations stem from environmental dynamics (objects displaced by humans), perceptual errors (incorrect initial state estimation), and manipulation failures (gripper slip, object drops).

Fig.~\ref{fig:situation_distribution} presents the distribution of situations across tasks. Out of 100 trials, 72 included at least one situation requiring detection and recovery. T2 (Cut Lemon) exhibited the highest situation rate due to the multi-step nature of the task and frequent scene disturbances during execution.

\begin{figure}[t]
    \centering
    \includegraphics[width=0.85\columnwidth]{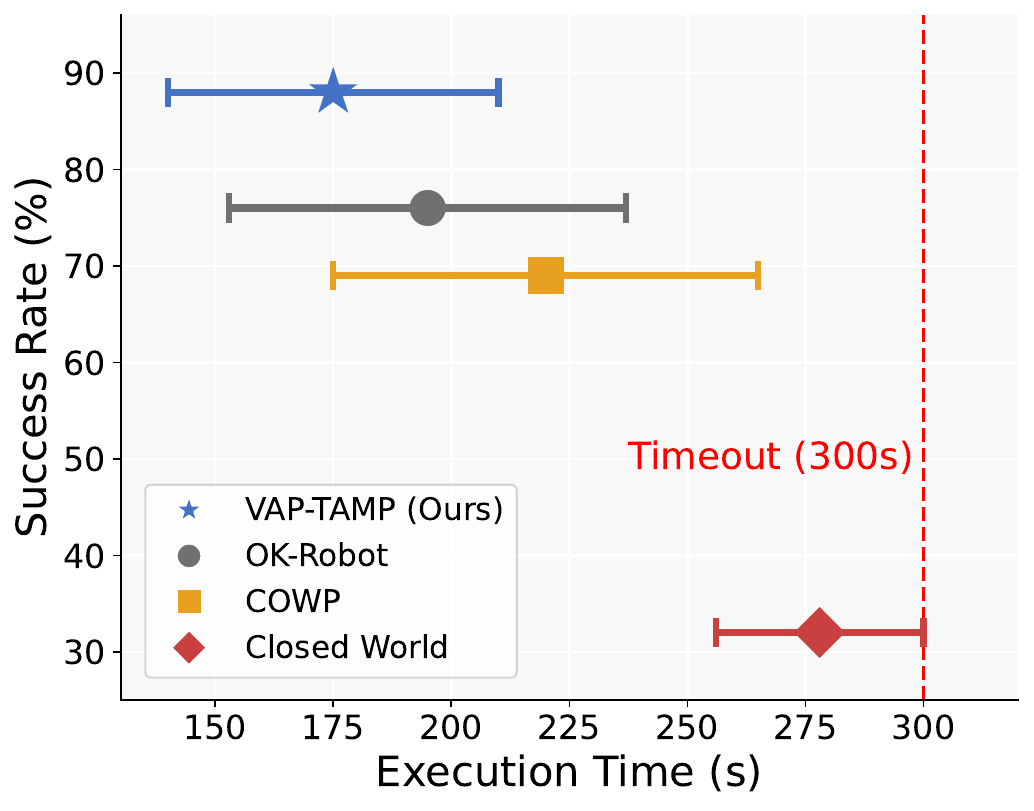}
    \vspace{-1.2em}
    \caption{Success rate vs. execution time across all methods. Points closer to the top-left indicate better overall performance. VAP-TAMP achieves the highest success rate with the lowest execution time.}
    \label{fig:time_efficiency}
\end{figure}

\begin{figure}[t]
    \centering
    \includegraphics[width=\columnwidth]{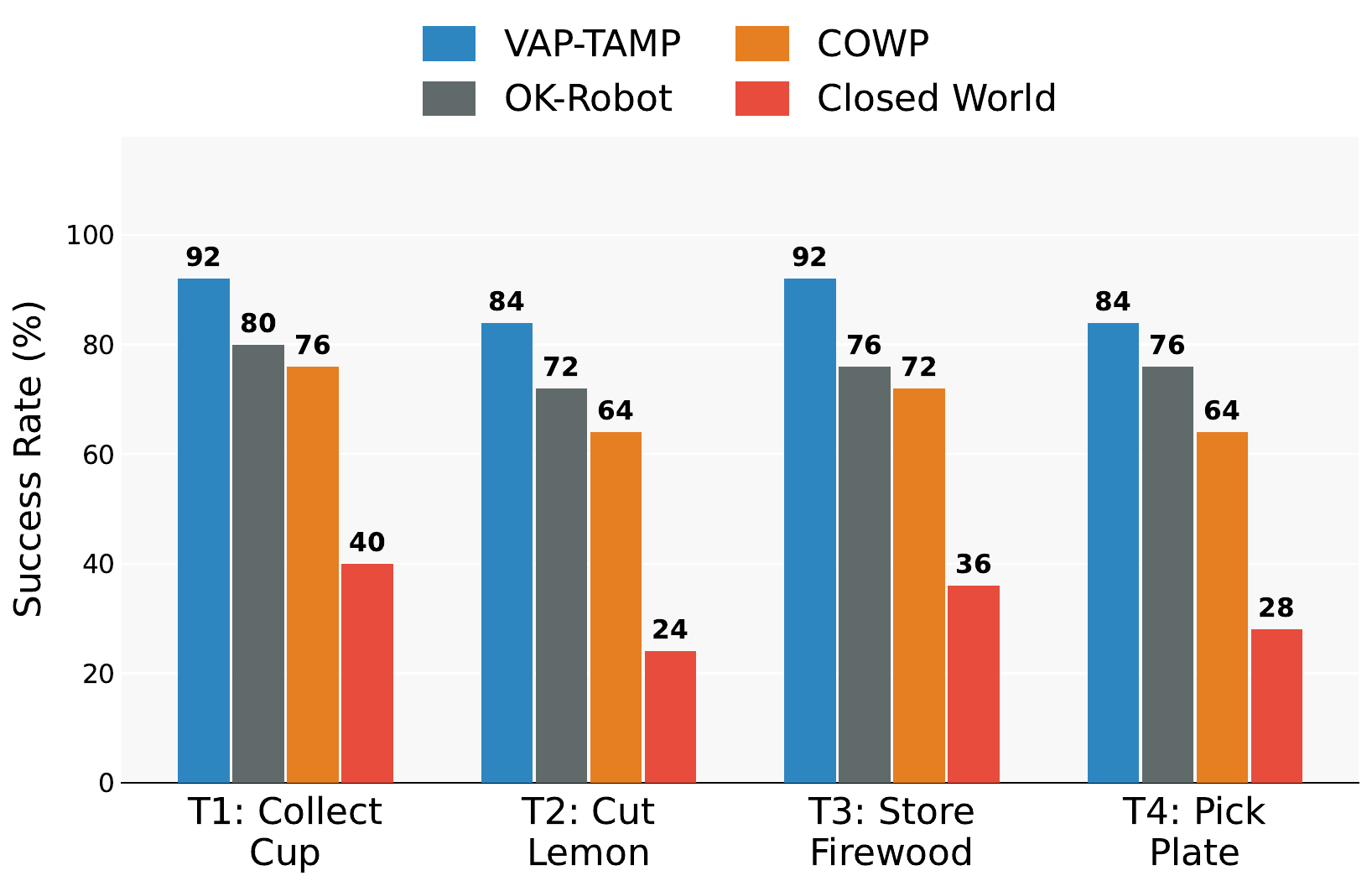}
     \vspace{-2.0em}
    \caption{Success rates by task. VAP-TAMP maintains consistent performance across all tasks, while baselines show larger variance.}
    \label{fig:success_by_task}
\end{figure}

\subsubsection{Results and Analysis}

\begin{figure}[t]
    \centering
    \includegraphics[width=0.98\columnwidth]{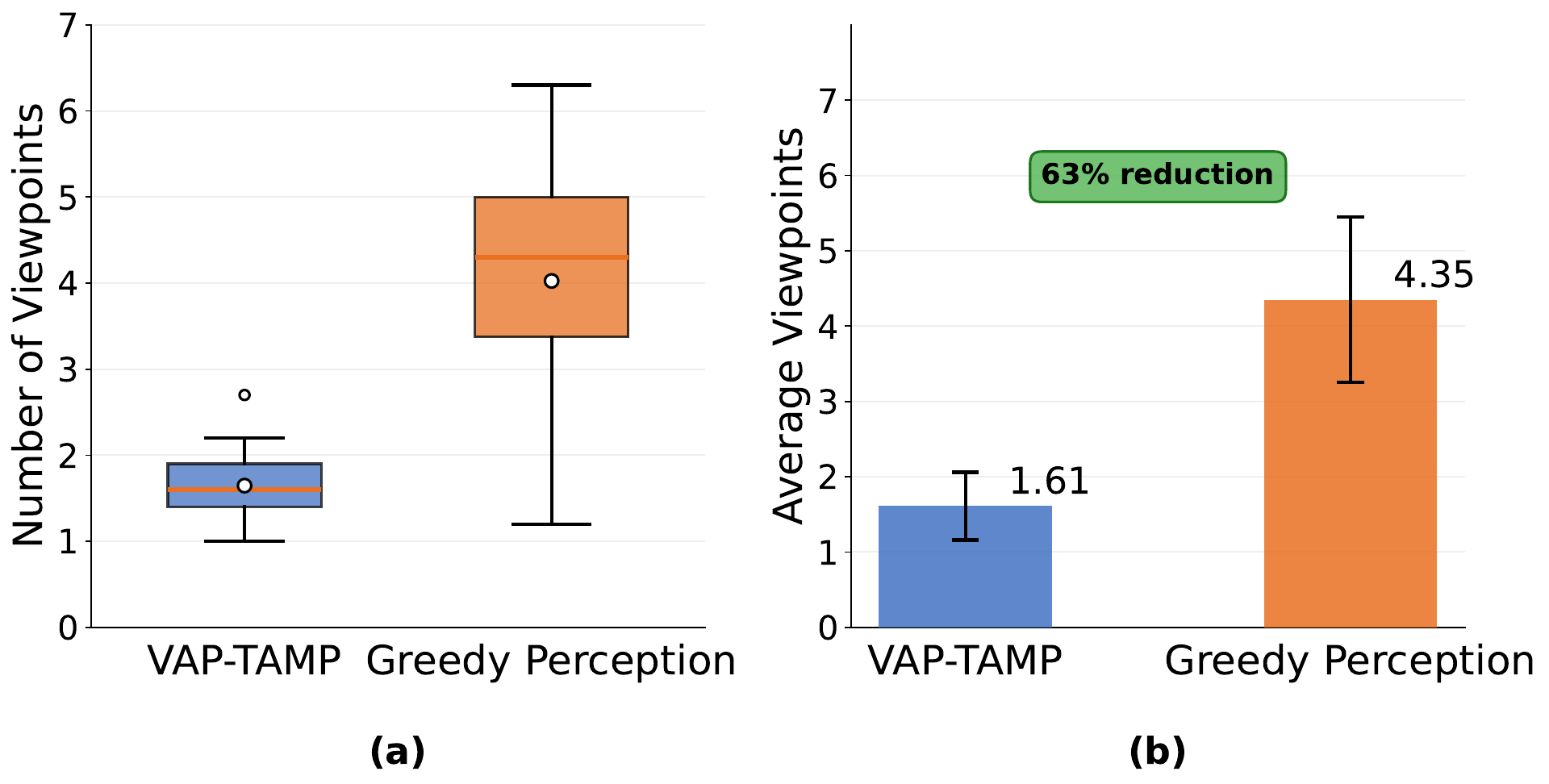}
     \vspace{-1.1em}
    \caption{Viewpoint efficiency: (a) distribution of viewpoints required per uncertainty resolution, (b) average viewpoints with standard deviation.}
    \label{fig:viewpoints}
\end{figure}

Fig.~\ref{fig:time_efficiency} shows results supporting H1. VAP-TAMP's improvement over OK-Robot demonstrates that active perception provides meaningful gains in predicate verification reliability. COWP's lower performance reflects the limitation of task planning without motion planning. Closed World's poor performance confirms that situation handling is essential in open-world environments.

Fig.~\ref{fig:success_by_task} shows per-task results. The largest performance gap occurs on T2, which has the highest situation rate. Scene disturbances during execution remain undetected by baselines, causing cascading failures. VAP-TAMP's situation handling detects state changes through precondition and effect verification, triggering replanning from the corrected state.

To evaluate H2, we compare VLM-guided viewpoint selection against a greedy baseline within VAP-TAMP. \textit{Greedy perception} exhaustively visits predefined viewpoints arranged around the target object. The number of predefined viewpoints varies by task (i.e. some tasks like Collect Cup require a range of 2-4 viewpoints (grasping within a confined 120° workspace), while multi-object tasks like Cut Lemon require 6 viewpoints (verifying knife,lemon, and cutting surface across 270°). Across all tasks, greedy perception averages 4.35 viewpoints. In contrast, \textit{VLM-guided perception} queries the VLM to suggest which direction would provide clearer evidence, navigating directly to discriminative viewpoints.
Fig.~\ref{fig:viewpoints} shows that VLM-guided selection resolves uncertainty over greedy perception. By reasoning about visual informativeness, VAP-TAMP avoids exhaustive exploration and navigates directly to disambiguating perspectives.

\begin{figure}[t]
    \centering
    \includegraphics[width=\columnwidth]{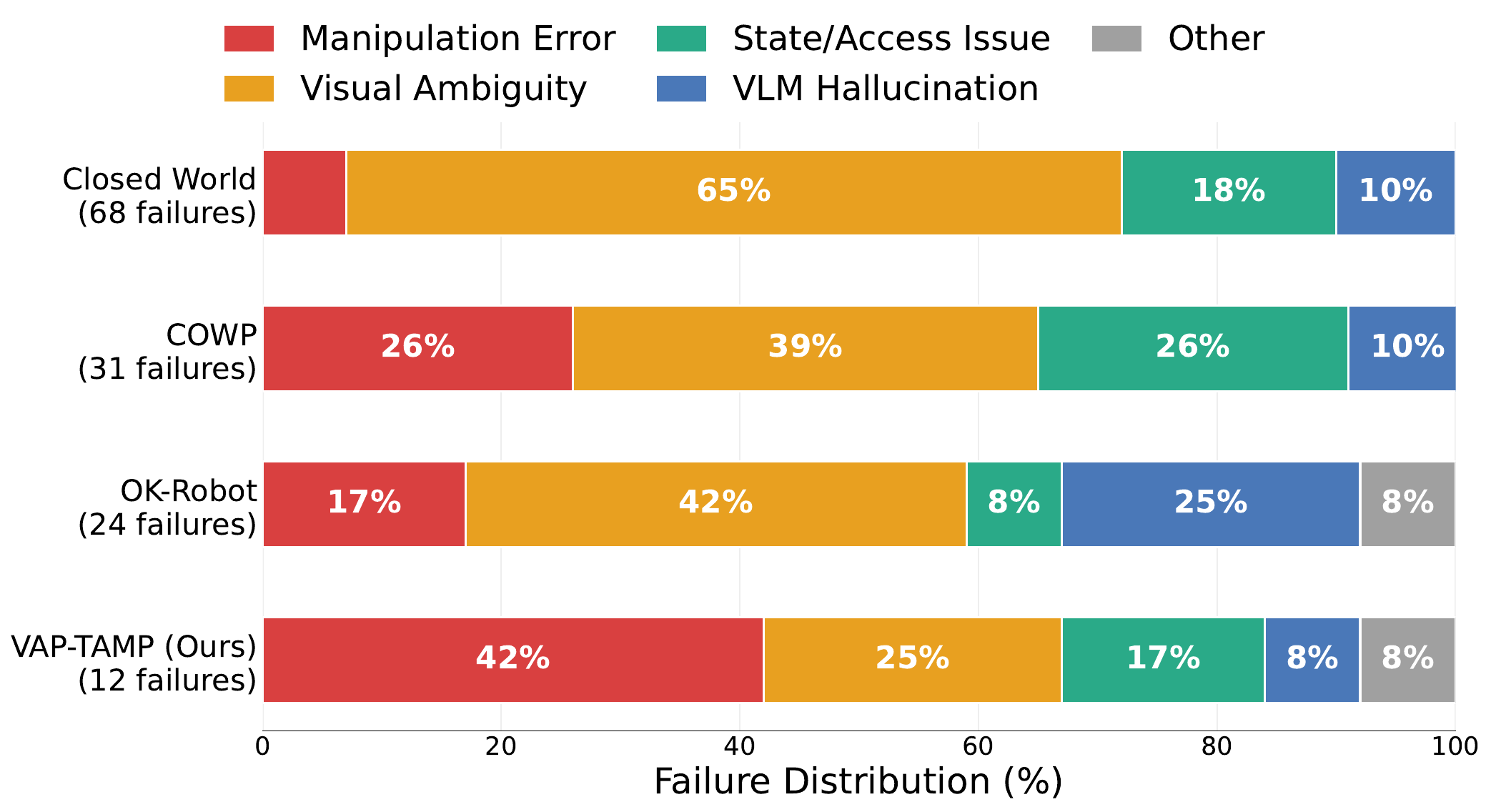}
    \vspace{-2.0em}
    \caption{Failure mode distribution across all methods.}
    \label{fig:failure_modes}
\end{figure}

Fig.~\ref{fig:failure_modes} reveals failure mode distributions across all methods. VAP-TAMP's failures are dominated by manipulation errors (hardware limitations), whereas OK-Robot and COWP show higher rates of visual ambiguity and state/access issues. Closed World's failures are dominated by visual ambiguity. By integrating active perception with situation handling, VAP-TAMP shifts the reliability bottleneck from perception to hardware: failures occur where the system lacks physical capability, not where it lacks perceptual information.

\subsection{Simulation Experiments}
\label{sec:simulation}

\subsubsection{Setup}

We use OmniGibson~\cite{li2023behavior} for simulation experiments, which provides realistic physics and diverse household environments. 
In simulation experiment, we focus on evaluating how PDDL-based action knowledge helps VLM prompting for actively evaluating pre- and post-conditions, while the robot's view point does not change with respect to the robot base. 
Simulation enables systematic evaluation of situation handling through controlled situation injection with predefined failure probabilities. 

We evaluate on five tasks with varying complexity. \textbf{S1: Boil Water} requires picking up an empty cup from a closed cabinet, filling it with water using a sink, and heating it in a microwave. \textbf{S2: Retrieve Bottles} requires finding two empty bottles in the garden and bringing them inside. \textbf{S3: Heat Frozen Pie} requires taking an apple pie out of the fridge and heating it using an oven. \textbf{S4: Halve Egg} requires finding a knife in the kitchen and using it to cut a hard-boiled egg in half. \textbf{S5: Gather Kindling} requires collecting two wooden sticks and placing them on a table. These tasks differ from real-world experiments to test generalization across diverse action sequences and object interactions not present in the physical setup.

We inject situations by assigning failure probabilities to each action type, as shown in Table~\ref{tab:sim_situations}. 
High manipulation actions have higher failure rates: \texttt{grasp} and \texttt{cut} each have 50\% combined failure probability (25\% object unchanged, 25\% object drops nearby). Simpler actions such as \texttt{open}, \texttt{close}, and \texttt{turnon} have 10\% failure rates. Some situations depend on external factors like navigation outcomes and are not directly injected.

\begin{table}[h]
    \centering
    \footnotesize
    \caption{Situation injection probabilities for simulation experiments. Each probability defines the likelihood of the corresponding situation occurring when the action is executed.}
    \vspace{0.5em}
    \begin{tabular}{llc}
        \toprule
        \textbf{Action} & \textbf{Situation} & \textbf{Prob.} \\
        \midrule
        \multirow{3}{*}{find} & Object not inview after navigation & N/A \\
        & No free space near object & N/A \\
        & Held object drops during navigation & 0.10 \\
        \midrule
        \multirow{2}{*}{grasp} & Grasp fails, object unchanged & 0.25 \\
        & Grasp fails, object drops nearby & 0.25 \\
        \midrule
        \multirow{2}{*}{placein} & Place fails, object remains inhand & 0.10 \\
        & Place fails, object drops nearby & 0.10 \\
        \midrule
        \multirow{2}{*}{placeon} & Place fails, object remains inhand & 0.10 \\
        & Place fails, object drops nearby & 0.10 \\
        \midrule
        \multirow{2}{*}{fill} & Container not fully filled & 0.05 \\
        & Container drops nearby & 0.05 \\
        \midrule
        open & Object remains closed & 0.10 \\
        \midrule
        close & Object remains open & 0.10 \\
        \midrule
        turnon & Object remains off & 0.10 \\
        \midrule
        \multirow{2}{*}{cut} & Object not cut, knife inhand & 0.25 \\
        & Object not cut, knife drops nearby & 0.25 \\
        \bottomrule
    \end{tabular}
    \label{tab:sim_situations}
\end{table}

We compare against verification strategies from the literature that differ in how they query the VLM for situation detection. \textbf{VLM-planner} uses the VLM to generate task plans directly, similar to SayCan~\cite{ahn2022can}; we provide domain knowledge in prompts for fair comparison. \textbf{Classical-planner} assumes all executions succeed without perception. \textbf{SuccessVQA} queries action success after execution (``Did the robot successfully \texttt{<action>}?''), inspired by~\cite{du2023vision}, but does not verify next-action executability. \textbf{AffordanceVQA} queries feasibility before execution (``Is it possible to \texttt{<action>}?''), based on prompts from PaLM-E~\cite{driess2023palm}, but does not verify previous-action success. \textbf{Suc.Aff.-QA} combines both strategies. Our predicate-based approach differs by querying concrete state predicates (e.g., ``Is the robot holding the cup?'') rather than abstract action outcomes.

\begin{figure}[t]
    \centering
    \includegraphics[width=\columnwidth]{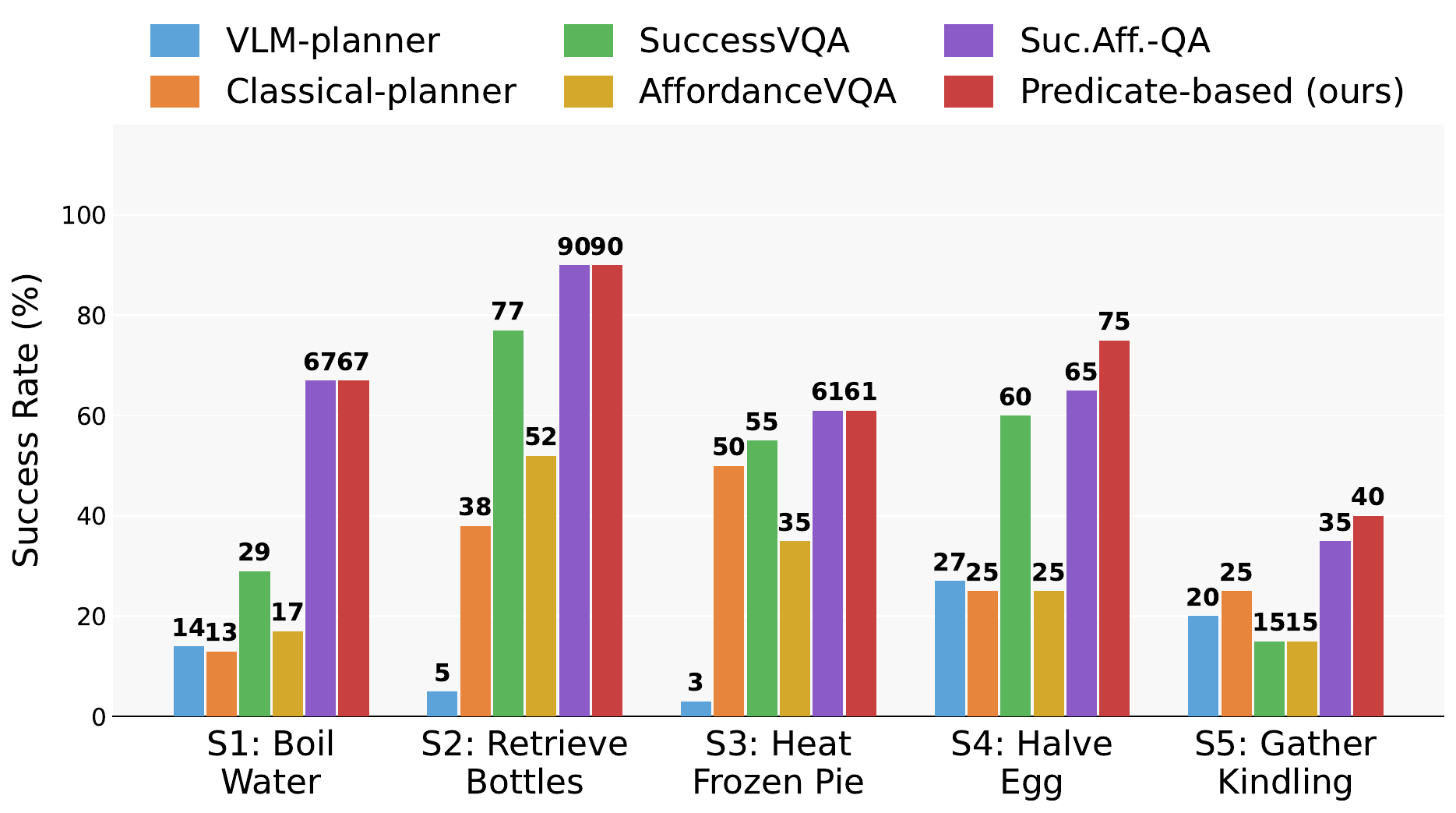}
    \vspace{-2.0em}
    \caption{Situation handling evaluation in simulation comparing verification strategies.}
    \label{fig:sim_baselines}
\end{figure}

\begin{table}[t]
    \centering
    \footnotesize
    \caption{Ablation of verification stages in simulation experiments.}
    \vspace{0.5em}
    \begin{tabular}{lcccccr}
        \toprule
        Verification & S1 & S2 & S3 & S4 & S5 & Avg. \\
        \midrule
        Both (full) & \textbf{66.7} & 90.0 & \textbf{60.9} & \textbf{75.0} & \textbf{40.0} & \textbf{66.5} \\
        Effects only & 50.0 & \textbf{93.8} & 26.7 & 66.7 & 28.0 & 53.0 \\
        Preconditions only & 17.6 & 75.0 & 35.0 & 55.0 & 20.0 & 41.5 \\
        \bottomrule
    \end{tabular}
    \label{tab:sim_ablation}
\end{table}

\subsubsection{Results}

Fig.~\ref{fig:sim_baselines} compares verification strategies. Our predicate-based approach outperforms all baselines across tasks. VLM-planner and Classical-planner lack verification entirely, leading to poor recovery from injected failures. SuccessVQA and AffordanceVQA each capture only one verification stage, missing failures detectable by the other. Even Suc.Aff.-QA, which combines both, underperforms our approach because VLMs reason more reliably about observable state predicates than abstract action outcomes.

Table~\ref{tab:sim_ablation} ablates verification stages. Both precondition and effect verification contribute to overall performance, with effect verification providing larger gains. This aligns with our situation injection: high-probability failures like \texttt{grasp} and \texttt{cut} (50\% combined failure rate each) are only detectable after execution through effect verification.

\section{Conclusion and Future Work}
\label{sec:conclusion}

We introduced VAP-TAMP, a task and motion planning framework that enables robots to actively perceive and address unforeseen situations during plan execution. 
Unlike prior approaches that assume situations are fully observable, VAP-TAMP handles partial observability by leveraging action knowledge to strategically prompt VLMs for situation assessment and active viewpoint selection. 
The framework closes the loop between planning and execution: when expected action conditions are not satisfied, the robot gathers disambiguating observations and replans with updated world state. 
Experimental results demonstrate substantial improvement over baselines, confirming that situation-aware planning with active perception is essential for reliable operation in open-world environments.

One direction for future work is to perform whole-body optimization, where currently the base movements (navigation) and upper body movements (manipulation) are optimized separately. 
Another direction is to integrate reinforcement learning to enable the robot to improve from failures. 
Finally, VAP-TAMP can be further strengthened to perform interactive perception for situation handling by physically changing the real-world configurations for perception.

\addtolength{\textheight}{-1cm}

\bibliographystyle{IEEEtran}
\bibliography{references}

\begin{thebibliography}{10}
\providecommand{\url}[1]{#1}
\csname url@samestyle\endcsname
\providecommand{\newblock}{\relax}
\providecommand{\bibinfo}[2]{#2}
\providecommand{\BIBentrySTDinterwordspacing}{\spaceskip=0pt\relax}
\providecommand{\BIBentryALTinterwordstretchfactor}{4}
\providecommand{\BIBentryALTinterwordspacing}{\spaceskip=\fontdimen2\font plus
\BIBentryALTinterwordstretchfactor\fontdimen3\font minus
  \fontdimen4\font\relax}
\providecommand{\BIBforeignlanguage}[2]{{%
\expandafter\ifx\csname l@#1\endcsname\relax
\typeout{** WARNING: IEEEtran.bst: No hyphenation pattern has been}%
\typeout{** loaded for the language `#1'. Using the pattern for}%
\typeout{** the default language instead.}%
\else
\language=\csname l@#1\endcsname
\fi
#2}}
\providecommand{\BIBdecl}{\relax}
\BIBdecl

\bibitem{garrett2021integrated}
C.~R. Garrett, R.~Chitnis, R.~Holladay, B.~Kim, T.~Silver, L.~P. Kaelbling, and
  T.~Lozano-P{\'e}rez, ``Integrated task and motion planning,'' \emph{Annual
  Review of Control, Robotics, and Autonomous Systems}, 2021.

\bibitem{zhao2024survey}
Z.~Zhao, S.~Cheng, Y.~Ding, Z.~Zhou, S.~Zhang, D.~Xu, and Y.~Zhao, ``A survey
  of optimization-based task and motion planning: From classical to learning
  approaches,'' \emph{IEEE/ASME Transactions on Mechatronics}, 2024.

\bibitem{hanheide2017robot}
M.~Hanheide, M.~G{\"o}belbecker, G.~S. Horn \emph{et~al.}, ``Robot task
  planning and explanation in open and uncertain worlds,'' \emph{Artificial
  Intelligence}, 2017.

\bibitem{ding2023integrating}
Y.~Ding, X.~Zhang, S.~Amiri, N.~Cao, H.~Yang, A.~Kaminski, C.~Esselink, and
  S.~Zhang, ``Integrating action knowledge and llms for task planning and
  situation handling in open worlds,'' \emph{Autonomous Robots}, 2023.

\bibitem{ren2023robots}
A.~Z. Ren, A.~Dixit, A.~Bodrova, S.~Singh, S.~Tu \emph{et~al.}, ``Robots that
  ask for help: Uncertainty alignment for large language model planners,''
  \emph{arXiv preprint arXiv:2307.01928}, 2023.

\bibitem{team2023gemini}
G.~Gemini~Team, ``Gemini: A family of highly capable multimodal models,''
  \emph{arXiv preprint arXiv:2312.11805}, 2023.

\bibitem{haslum2019introduction}
P.~Haslum, N.~Lipovetzky, D.~Magazzeni, C.~Muise, R.~Brachman, F.~Rossi, and
  P.~Stone, \emph{An introduction to the planning domain definition
  language}.\hskip 1em plus 0.5em minus 0.4em\relax Springer, 2019, vol.~13.

\bibitem{Liu_2024}
P.~Liu, Y.~Orru, J.~Vakil, C.~Paxton, N.~Shafiullah, and L.~Pinto, ``Ok-robot:
  What really matters in integrating open-knowledge models for robotics,'' in
  \emph{Robotics: Science and Systems (RSS)}, 2024.

\bibitem{kaelbling2013integrated}
L.~P. Kaelbling and T.~Lozano-P{\'e}rez, ``Integrated task and motion planning
  in belief space,'' \emph{The International Journal of Robotics Research},
  2013.

\bibitem{garrett2020pddlstream}
C.~R. Garrett, T.~Lozano-P{\'e}rez, and L.~P. Kaelbling, ``Pddlstream:
  Integrating symbolic planners and blackbox samplers,'' in \emph{International
  Conference on Automated Planning and Scheduling (ICAPS)}, 2020.

\bibitem{garrett2018ffrob}
C.~R. Garrett, T.~Lozano-Perez, and L.~P. Kaelbling, ``Ffrob: Leveraging
  symbolic planning for efficient task and motion planning,'' \emph{The
  International Journal of Robotics Research}, 2018.

\bibitem{zhang2025llm}
X.~Zhang, Y.~Ding, Y.~Hayamizu, Z.~Altaweel, Y.~Zhu, Y.~Zhu, P.~Stone,
  C.~Paxton, and S.~Zhang, ``Llm-grop: Visually grounded robot task and motion
  planning with large language models,'' \emph{The International Journal of
  Robotics Research}, 2025.

\bibitem{agarwal2025l3m+}
K.~Agarwal, Y.~Jiang, J.~Hu, B.~Liu, and P.~Stone, ``L3m+p: Lifelong planning
  with large language models,'' in \emph{IEEE/RSJ International Conference on
  Intelligent Robots and Systems (IROS)}, 2025.

\bibitem{curtis2024partially}
A.~Curtis, G.~Matheos, N.~Gothoskar, V.~Mansinghka, J.~Tenenbaum,
  T.~Lozano-P{\'e}rez, and L.~P. Kaelbling, ``Partially observable task and
  motion planning with uncertainty and risk awareness,'' \emph{arXiv preprint
  arXiv:2403.10454}, 2024.

\bibitem{chitnis2016guided}
R.~Chitnis, D.~Hadfield-Menell, A.~Gupta, S.~Srivastava, E.~Groshev, C.~Lin,
  and P.~Abbeel, ``Guided search for task and motion plans using learned
  heuristics,'' in \emph{2016 IEEE International Conference on Robotics and
  Automation (ICRA)}.\hskip 1em plus 0.5em minus 0.4em\relax IEEE, 2016, pp.
  447--454.

\bibitem{radford2021learning}
A.~Radford, J.~W. Kim, C.~Hallacy, A.~Ramesh, G.~Goh \emph{et~al.}, ``Learning
  transferable visual models from natural language supervision,'' in
  \emph{International Conference on Machine Learning (ICML)}, 2021.

\bibitem{ahn2022can}
M.~Ahn, A.~Brohan, N.~Brown, Y.~Chebotar, O.~Cortes, B.~David, C.~Finn
  \emph{et~al.}, ``Do as i can, not as i say: Grounding language in robotic
  affordances,'' \emph{arXiv preprint arXiv:2204.01691}, 2022.

\bibitem{liang2022code}
J.~Liang, W.~Huang, F.~Xia, P.~Xu, K.~Hausman \emph{et~al.}, ``Code as
  policies: Language model programs for embodied control,'' \emph{arXiv
  preprint arXiv:2209.07753}, 2022.

\bibitem{driess2023palm}
D.~Driess, F.~Xia, M.~S. Sajjadi, C.~Lynch, A.~Chowdhery \emph{et~al.},
  ``Palm-e: An embodied multimodal language model,'' 2023.

\bibitem{huang2023voxposer}
W.~Huang, C.~Wang, R.~Zhang, Y.~Li, J.~Wu, and L.~Fei-Fei, ``Voxposer:
  Composable 3d value maps for robotic manipulation with language models,''
  \emph{arXiv preprint arXiv:2307.05973}, 2023.

\bibitem{rahmanzadehgervi2024vision}
P.~Rahmanzadehgervi, L.~Bolton, M.~R. Taesiri, and A.~T. Nguyen, ``Vision
  language models are blind,'' \emph{arXiv preprint arXiv:2407.06581}, 2024.

\bibitem{lin2023generating}
Z.~Lin, S.~Trivedi, and J.~Sun, ``Generating with confidence: Uncertainty
  quantification for black-box large language models,'' \emph{arXiv preprint
  arXiv:2305.19187}, 2023.

\bibitem{bajcsy1988active}
R.~Bajcsy, ``Active perception,'' \emph{Proceedings of the IEEE}, 1988.

\bibitem{bajcsy2018revisiting}
R.~Bajcsy, Y.~Aloimonos, and J.~K. Tsotsos, ``Revisiting active perception,''
  \emph{Autonomous Robots}, 2018.

\bibitem{kriegel2011surface}
S.~Kriegel, T.~Bodenm{\"u}ller, M.~Suppa, and G.~Hirzinger, ``A surface-based
  next-best-view approach for automated 3d model completion of unknown
  objects,'' in \emph{IEEE International Conference on Robotics and Automation
  (ICRA)}, 2011.

\bibitem{isler2016information}
S.~Isler, R.~Sabzevari, J.~Delmerico, and D.~Scaramuzza, ``An information gain
  formulation for active volumetric 3d reconstruction,'' in \emph{IEEE
  International Conference on Robotics and Automation (ICRA)}, 2016.

\bibitem{bohg2017interactive}
J.~Bohg, K.~Hausman, B.~Sankaran, O.~Brock, D.~Kragic, S.~Schaal, and G.~S.
  Sukhatme, ``Interactive perception: Leveraging action in perception and
  perception in action,'' \emph{IEEE Transactions on Robotics}, 2017.

\bibitem{zhang2021planning}
X.~Zhang, J.~Sinapov, and S.~Zhang, ``Planning multimodal exploratory actions
  for online robot attribute learning,'' in \emph{Proceedings of Robotics:
  Science and Systems}, 2021.

\bibitem{rosinol2021kimera}
A.~Rosinol, A.~Violette, M.~Abate, N.~Hughes, Y.~Chang \emph{et~al.}, ``Kimera:
  From slam to spatial perception with 3d dynamic scene graphs,'' \emph{The
  International Journal of Robotics Research}, 2021.

\bibitem{hughes2022hydra}
N.~Hughes, Y.~Chang, and L.~Carlone, ``Hydra: A real-time spatial perception
  system for 3d scene graph construction and optimization,'' \emph{arXiv
  preprint arXiv:2201.13360}, 2022.

\bibitem{amiri2022reasoning}
S.~Amiri, S.~Wei, S.~Zhang, J.~Sinapov, and P.~Stone, ``Reasoning with scene
  graphs for robot planning under partial observability,'' \emph{IEEE Robotics
  and Automation Letters}, 2022.

\bibitem{gu2024conceptgraphs}
Q.~Gu, A.~Kuwajerwala, S.~Morin, K.~M. Jatavallabhula, B.~Sen \emph{et~al.},
  ``Conceptgraphs: Open-vocabulary 3d scene graphs for perception and
  planning,'' in \emph{IEEE International Conference on Robotics and Automation
  (ICRA)}, 2024.

\bibitem{ingrand2017deliberation}
F.~Ingrand and M.~Ghallab, ``Deliberation for autonomous robots: A survey,''
  \emph{Artificial Intelligence}, 2017.

\bibitem{brenner2006continual}
M.~Brenner and B.~Nebel, ``Continual planning and acting in dynamic multiagent
  environments,'' in \emph{International Symposium on Practical Cognitive
  Agents and Robots}, 2006.

\bibitem{du2023vision}
Y.~Du, K.~Konyushkova, M.~Denil, A.~Raju \emph{et~al.}, ``Vision-language
  models as success detectors,'' \emph{arXiv preprint arXiv:2303.07280}, 2023.

\bibitem{huang2022inner}
W.~Huang, F.~Xia, T.~Xiao, H.~Chan, J.~Liang \emph{et~al.}, ``Inner monologue:
  Embodied reasoning through planning with language models,'' \emph{arXiv
  preprint arXiv:2207.05608}, 2022.

\bibitem{rana2023sayplan}
K.~Rana, J.~Haviland, S.~Garg, J.~Abou-Chakra, I.~Reid, and N.~Suenderhauf,
  ``Sayplan: Grounding large language models using 3d scene graphs for scalable
  robot task planning,'' \emph{arXiv preprint arXiv:2307.06135}, 2023.

\bibitem{coleman2014reducing}
D.~Coleman, I.~Sucan, S.~Chitta, and N.~Correll, ``Reducing the barrier to
  entry of complex robotic software: a moveit! case study,'' in \emph{Journal
  of Software Engineering for Robotics}, 2014.

\bibitem{li2023behavior}
C.~Li, R.~Zhang, J.~Wong, C.~Gokmen, S.~Srivastava \emph{et~al.},
  ``Behavior-1k: A benchmark for embodied ai with 1,000 everyday activities and
  realistic simulation,'' in \emph{Conference on Robot Learning (CoRL)}, 2023.

\end{thebibliography}

\clearpage
\begin{appendices}

\section{VLM Prompt Templates}
\label{sec:prompts}

This appendix provides the complete set of prompt templates used for VLM queries in VAP-TAMP. As described in Section~\ref{sec:methodology}, the system uses VLMs for: (1) parsing natural language goals into PDDL specifications, (2) verifying predicates through paraphrased queries, (3) assessing view sufficiency, and (4) suggesting better viewpoints when needed. All prompts are designed to elicit binary or structured responses for reliable automated parsing.

\subsubsection{Goal Parsing}
\label{appendix:goal_parsing}

Given a natural language instruction $g_{\text{NL}}$, the system uses the VLM to generate a PDDL goal specification $g$ based on the current scene graph $\mathcal{G}$ and domain knowledge $\mathcal{D}$.

\begin{tcolorbox}[colback=gray!5, colframe=gray!50, title=Goal Parsing Prompt]
\small
\texttt{You are a robot task planner. Convert the natural language instruction into a goal state.}

\vspace{0.5em}
\texttt{Instruction: \{instruction\}}

\vspace{0.5em}
\texttt{Available objects: \{object\_list\}}

\vspace{0.5em}
\texttt{Available predicates: \{predicate\_list\}}

\vspace{0.5em}
\texttt{Output the goal as a list of predicates that should be true when the task is complete.}

\vspace{0.5em}
\texttt{Format: predicate(object1, object2), predicate(object), ...}

\vspace{0.5em}
\texttt{Goal:}

\vspace{0.5em}
\texttt{Example:}
\begin{itemize}[leftmargin=*, nosep]
    \item \texttt{Input: Instruction = ``Put the cup in the cabinet''}
    \item \texttt{Objects = [cup, cabinet, table, plate]}
    \item \texttt{Predicates = [on(X,Y), inside(X,Y), holding(X), open(X)]}
    \item \texttt{Output: inside(cup, cabinet)}
\end{itemize}
\end{tcolorbox}

\subsubsection{Predicate Verification}
\label{appendix:predicate_verification}

As described in Section~\ref{sec:active_perception}, predicate verification uses $N=5$ semantically equivalent paraphrases to improve reliability. Each paraphrase is sent to the VLM with the current observation, and the final predicate value is determined by majority vote over the responses.

\begin{tcolorbox}[colback=gray!5, colframe=gray!50, title=Predicate Query Template]
\small
\texttt{[Image attached]}

\vspace{0.5em}
\texttt{Analyze this image carefully.}

\vspace{0.5em}
\texttt{Question: \{paraphrased\_question\}}

\vspace{0.5em}
\texttt{Respond with only ``yes'' or ``no''.}
\end{tcolorbox}

For example, to verify predicate $p = \texttt{on(cup, table)}$, we generate five paraphrased questions and query the VLM five times with the same observation. If 4 out of 5 responses are ``yes'', the predicate evaluates to True. Section~\ref{appendix:paraphrases} provides the complete set of paraphrase templates for all predicates.

\subsubsection{View Sufficiency Check}
\label{appendix:view_sufficiency}

When predicate verification produces consistent responses (high agreement among paraphrases), the system queries the VLM to determine whether the current viewpoint provides adequate visual information to confidently accept the result. If the VLM indicates insufficiency, viewpoint selection is triggered.

\begin{tcolorbox}[colback=gray!5, colframe=gray!50, title=View Sufficiency Prompt]
\small
\texttt{[Image attached]}

\vspace{0.5em}
\texttt{You are assessing whether this camera view provides sufficient information to answer the following question:}

\vspace{0.5em}
\texttt{``\{predicate\_question\}''}

\vspace{0.5em}
\texttt{Consider:}
\begin{itemize}[leftmargin=*, nosep]
    \item \texttt{Is the target object clearly visible?}
    \item \texttt{Are relevant spatial relationships observable?}
    \item \texttt{Is the view free from significant occlusion?}
\end{itemize}

\vspace{0.5em}
\texttt{Respond ``yes'' if the current view is sufficient.}

\texttt{Respond ``no'' if a different viewpoint would provide clearer evidence.}

\vspace{0.5em}
\texttt{Answer:}
\end{tcolorbox}

\subsubsection{Viewpoint Selection}
\label{appendix:viewpoint_selection}

When the view is insufficient, the system queries the VLM to suggest a better viewing direction. The robot navigates in the suggested direction and re-attempts predicate verification, repeating until the view is sufficient or the viewpoint budget $K$ is exhausted.

\begin{tcolorbox}[colback=gray!5, colframe=gray!50, title=Viewpoint Selection Prompt]
\small
\texttt{[Image attached]}

\vspace{0.5em}
\texttt{The robot is trying to verify: ``\{predicate\_question\}''}

\vspace{0.5em}
\texttt{The current view does not provide sufficient visual evidence. Suggest which direction the robot should move to get a clearer view of the \{target\_object\}.}

\vspace{0.5em}
\texttt{Options: left, right, front, behind, above, closer}

\vspace{0.5em}
\texttt{Choose the single best direction:}
\end{tcolorbox}

\subsubsection{Predicate Paraphrase Templates}
\label{appendix:paraphrases}

For each predicate type in the domain, we define $N=5$ semantically equivalent phrasings. These paraphrases reduce sensitivity to specific wording and enable majority voting for robust verification.

\subsubsection{on(X, Y) -- Object Resting on Surface}

\begin{tcolorbox}[colback=blue!5, colframe=blue!30]
\small
\begin{enumerate}[leftmargin=*, nosep]
    \item \texttt{Is the \{X\} on the \{Y\}?}
    \item \texttt{Is the \{X\} resting on the \{Y\} surface?}
    \item \texttt{Is the \{X\} placed on top of the \{Y\}?}
    \item \texttt{Is the \{X\} positioned on the \{Y\}?}
    \item \texttt{Is the \{X\} sitting on the \{Y\}?}
\end{enumerate}
\end{tcolorbox}

\subsubsection{inside(X, Y) -- Containment}

\begin{tcolorbox}[colback=blue!5, colframe=blue!30]
\small
\begin{enumerate}[leftmargin=*, nosep]
    \item \texttt{Is the \{X\} inside the \{Y\}?}
    \item \texttt{Is the \{X\} contained within the \{Y\}?}
    \item \texttt{Can you see the \{X\} stored inside the \{Y\}?}
    \item \texttt{Is the \{X\} located within the \{Y\}?}
    \item \texttt{Is the \{X\} placed inside the \{Y\}?}
\end{enumerate}
\end{tcolorbox}

\subsubsection{holding(robot, X) -- Gripper Grasping Object}

\begin{tcolorbox}[colback=blue!5, colframe=blue!30]
\small
\begin{enumerate}[leftmargin=*, nosep]
    \item \texttt{Is the robot holding the \{X\}?}
    \item \texttt{Is the \{X\} grasped by the robot gripper?}
    \item \texttt{Does the robot have the \{X\} in its gripper?}
    \item \texttt{Is the robot's gripper gripping the \{X\}?}
    \item \texttt{Is the \{X\} held by the robot?}
\end{enumerate}
\end{tcolorbox}

\subsubsection{hand\_empty -- Empty Gripper}

\begin{tcolorbox}[colback=blue!5, colframe=blue!30]
\small
\begin{enumerate}[leftmargin=*, nosep]
    \item \texttt{Is the robot gripper empty?}
    \item \texttt{Is the robot holding nothing?}
    \item \texttt{Is the robot's gripper free and not grasping anything?}
    \item \texttt{Are the robot's fingers not holding any object?}
    \item \texttt{Is there nothing in the robot's gripper?}
\end{enumerate}
\end{tcolorbox}

\subsubsection{open(X) -- Door or Container State}

\begin{tcolorbox}[colback=blue!5, colframe=blue!30]
\small
\begin{enumerate}[leftmargin=*, nosep]
    \item \texttt{Is the \{X\} open?}
    \item \texttt{Is the \{X\} in an open position?}
    \item \texttt{Can you see inside the \{X\}, indicating it is open?}
    \item \texttt{Is the \{X\} door/lid currently opened?}
    \item \texttt{Is the interior of the \{X\} visible and accessible?}
\end{enumerate}
\end{tcolorbox}

\subsubsection{reachable(X) -- Object Accessibility}

\begin{tcolorbox}[colback=blue!5, colframe=blue!30]
\small
\begin{enumerate}[leftmargin=*, nosep]
    \item \texttt{Can the robot reach the \{X\}?}
    \item \texttt{Is the \{X\} accessible to the robot arm?}
    \item \texttt{Is there a clear path for the robot to reach the \{X\}?}
    \item \texttt{Can the robot arm access the \{X\} without obstruction?}
    \item \texttt{Is the \{X\} within the robot's reachable workspace?}
\end{enumerate}
\end{tcolorbox}

\subsubsection{blocking(X, Y) -- Obstruction}

\begin{tcolorbox}[colback=blue!5, colframe=blue!30]
\small
\begin{enumerate}[leftmargin=*, nosep]
    \item \texttt{Is the \{X\} blocking access to the \{Y\}?}
    \item \texttt{Is the \{X\} obstructing the \{Y\}?}
    \item \texttt{Does the \{X\} prevent reaching the \{Y\}?}
    \item \texttt{Is the \{X\} in the way of the \{Y\}?}
    \item \texttt{Would the \{X\} need to be moved to access the \{Y\}?}
\end{enumerate}
\end{tcolorbox}

\subsubsection{clear(X) -- Surface Unoccupied}

\begin{tcolorbox}[colback=blue!5, colframe=blue!30]
\small
\begin{enumerate}[leftmargin=*, nosep]
    \item \texttt{Is the top of the \{X\} clear?}
    \item \texttt{Is there nothing on top of the \{X\}?}
    \item \texttt{Is the \{X\} surface empty?}
    \item \texttt{Is the top of the \{X\} free of objects?}
    \item \texttt{Can an object be placed on the \{X\} without obstruction?}
\end{enumerate}
\end{tcolorbox}

\subsubsection{inview(X) -- Object Visibility}

\begin{tcolorbox}[colback=blue!5, colframe=blue!30]
\small
\begin{enumerate}[leftmargin=*, nosep]
    \item \texttt{Is the \{X\} visible in this image?}
    \item \texttt{Can you see the \{X\} in this view?}
    \item \texttt{Is the \{X\} present and visible in this image?}
    \item \texttt{Does this image contain the \{X\}?}
    \item \texttt{Is the \{X\} observable from this viewpoint?}
\end{enumerate}
\end{tcolorbox}
\end{appendices}
\end{document}